\documentclass[letterpaper, 10 pt, conference]{ieeeconf}  
\IEEEoverridecommandlockouts                              

\usepackage{graphicx} 
\usepackage{times} 
\usepackage{cite}
\usepackage{amsmath} 
\usepackage{amssymb}  
\usepackage{amsfonts}       
\usepackage{dsfont}
\usepackage{braket}
\usepackage{booktabs}
\usepackage{color}
\usepackage{colortbl}
\usepackage{url}
\usepackage{mathtools}
\usepackage{multirow}
\usepackage{footmisc}
\usepackage[linesnumbered,ruled,vlined]{algorithm2e}
\SetKwInput{KwInput}{Input}                
\SetKwInput{KwOutput}{Output}              

\usepackage[caption=false, font=footnotesize]{subfig}

\usepackage{lipsum}
\usepackage{widetext}

\newcommand{\optimal}{\mathcal{O}}

\newcommand{\traj}{\tau}
\newcommand{\obs}{\boldsymbol{x}}
\newcommand{\action}{\boldsymbol{a}}
\newcommand{\actionm}{\boldsymbol\mu}
\newcommand{\actions}{\boldsymbol\sigma}
\newcommand{\modelpost}{p(\theta|\mathcal{D})}

\newcommand{\latent}{{\boldsymbol z}}
\newcommand{\latents}{{\boldsymbol s}}
\newcommand{\latenth}{{\boldsymbol h}}

\newcommand{\mixcoef}{\pi}
\newcommand{\resp}{\eta}
\newcommand{\bhline}[1]{\noalign{\hrule height #1}}

\SetCommentSty{mycommfont}

\title{\LARGE \textbf{
PlaNet of the Bayesians: Reconsidering and Improving \\ Deep Planning Network
by Incorporating Bayesian Inference
}}

\author{Masashi Okada$^{\dag,\star}$, Norio Kosaka$^{\dag}$ and Tadahiro Taniguchi$^{\dag,*}$
\thanks{$^{\dag}$ Masashi Okada, Norio Kosaka, and Tadahiro Taniguchi are with AI Solutions Center, Business Innovation Division, Panasonic Corporation, Japan.
}%
\thanks{$^{*}$ Tadahiro Taniguchi is also with Ritsumeikan University, College of Information Science and Engineering, Japan.
}%
\thanks{$^{\star}$ Corresponding author: \texttt{okada.masashi001@jp.panasonic.com}
}
}

\begin{document}

\maketitle
\thispagestyle{empty}
\pagestyle{empty}

\begin{abstract}
In the present paper, we propose an extension of the Deep Planning Network (PlaNet), also referred to as PlaNet of the Bayesians (PlaNet-Bayes).
There has been a growing demand in model predictive control (MPC) in partially observable environments in which complete information is unavailable because of, for example, lack of expensive sensors.
PlaNet is a promising solution to realize such latent MPC, as it is used to train state-space models via model-based reinforcement learning (MBRL) and to conduct planning in the latent space.
However, recent state-of-the-art strategies mentioned in MBRR literature, such as involving uncertainty into training and planning, have not been considered, significantly suppressing the training performance.
The proposed extension is to make PlaNet uncertainty-aware on the basis of Bayesian inference, in which both model and action uncertainty are incorporated.
Uncertainty in latent models is represented using a neural network ensemble  to approximately infer model posteriors.
The ensemble of optimal action candidates is also employed to capture multimodal uncertainty in the optimality.
The concept of the action ensemble relies on a general variational inference MPC (VI-MPC) framework and its instance, probabilistic action ensemble with trajectory sampling (PaETS).
In this paper, we extend VI-MPC and PaETS, which have been originally introduced in previous literature, to address partially observable cases.
We experimentally compare the performances on continuous control tasks, and conclude that our method can consistently improve the asymptotic performance compared with PlaNet.
\end{abstract}

\section{Introduction} \label{sec:intro}
In the present paper, we focus on model predictive control (MPC) in partially observable environments.
MPC is a promising technique used in advanced control systems
that relies on the specified system models to predict future states and rewards for the purpose of planning.
The clear explainability of such decision-making processes is preferable, especially for industrial systems,
and therefore, many real-world applications have introduced MPC such as
HVAC systems~\cite{afram2014theory},
manufacturing processes~\cite{vargas2000multilayer}, 
and power electronics~\cite{vazquez2014model}.
The above systems assume fully observable environments; however, in practical applications, complete information is often unavailable, as measuring sufficient information for planning may be difficult and/or require expensive devices (for example, LiDARs).

\begin{figure}[tb]
  \centering
  \includegraphics[width=0.5\textwidth]{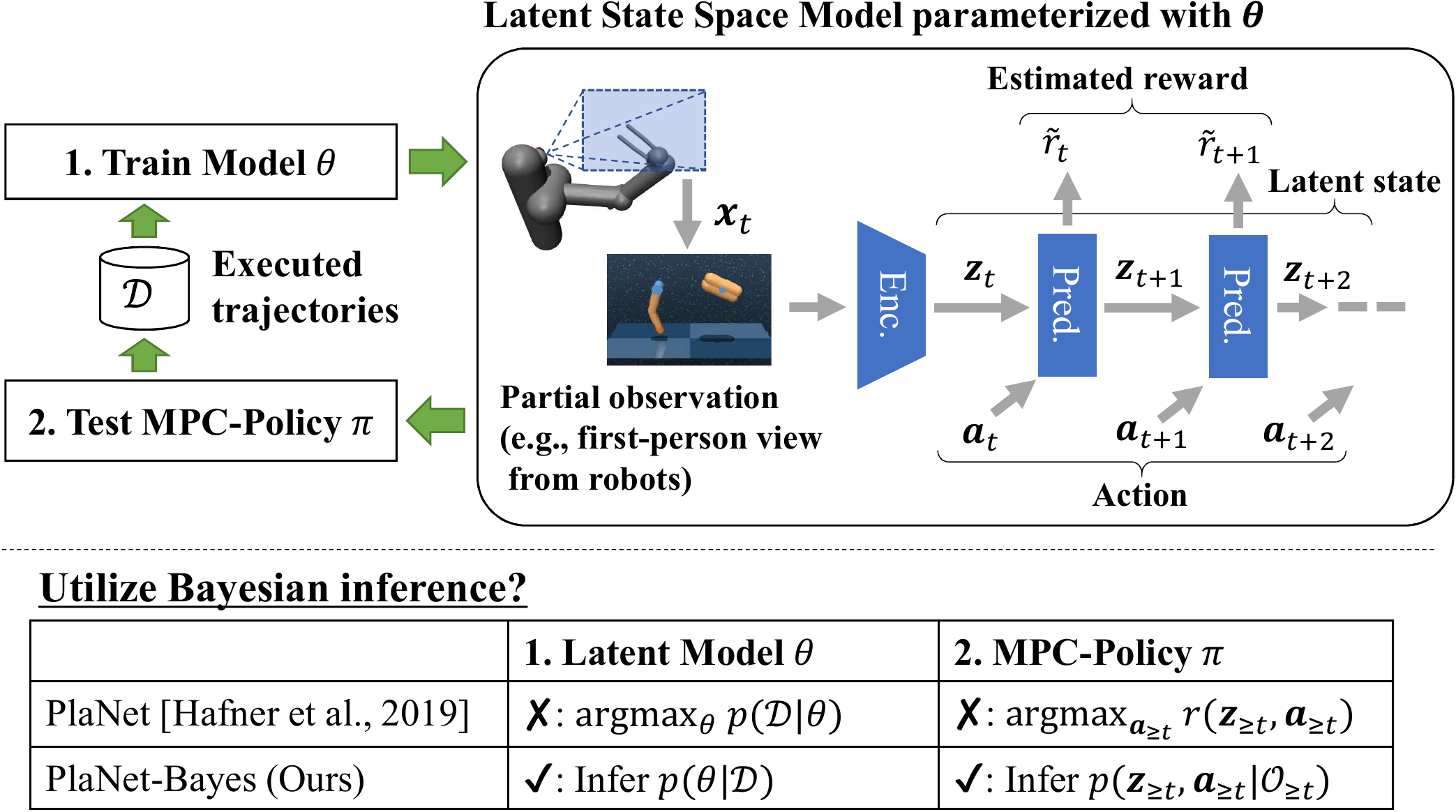}
  \caption{Top: the concept of latent state-space model training and planning in the learned latent space.
  Buttom: comparison of PlaNet~\cite{hafner2018learning} and the proposed PlaNet-Bayes.
  $\mathcal{O}$ indicates a random variable called \textit{optimality}, which is introduced later in Sec.~\ref{sec:vimpc}.
}
  \label{fig:fig1}
\end{figure}

The Deep Planning Network (PlaNet)~\cite{hafner2018learning} is an MPC-oriented model-based reinforcement learning (MBRL) method used for partially observable environments.
Fig.~\ref{fig:fig1} outlines the concept of PlaNet in which
an encoder is utilized to convert partial observations $\obs_{t}$ (e.g., high-dimensional raw observations such as images) into latent states $\latent_{t}$.
In the latent space, a prediction model is used to estimates next latent states $\latent_{t+1}$ and rewards $\tilde{r}_{t}$ based on current states $\latent_{t}$ and given actions $\action_{t}$.
This modeling way allows planning in the latent state-space in which we expect that the sufficient information for planning is embedded.
As our primary interest is to apply MPC to practical systems, in this paper, we mainly focus on PlaNet and consider this method as a strong baseline.

Recent studies on MBRL in fully observable environments have shown that uncertainty-aware modeling is essential to enhance the training performance.
The \textit{model-bias problem}~\cite{deisenroth2011pilco}, which occurs because of the overfitting of model parameters $\theta$ with respect to the limited amount of data $\mathcal{D}$ available during an early training phase, has been regarded as an inherent problem that restrains the MBRL potential.
However, recent studies have demonstrated that incorporating uncertainty in the model parameters can alleviate this issue by exploiting Bayesian inference of posterior $p(\theta|\mathcal{D})$%
~\cite{deisenroth2011pilco,gal2016dropout,gal2017concrete,kahn2017uncertainty,chua2018deep,kurutach2018model,clavera2018model,okada2019variational,nagabandi2019deep}.
In addition, in ~\cite{okada2019variational}, they have shown that involving uncertainty in optimal actions also affects performance improvement.
In the literature, the trajectory optimization problem in the Markov decision process is formulated as a variational inference problem~\cite{levine2018reinforcement}, deriving a general framework called variational inference MPC (VI-MPC).
As an instance of the framework, probabilistic action ensemble with trajectory sampling (PaETS) is also proposed that uses Gaussian mixture model (GMM) as the variational distribution by using which we can naturally model multimodal uncertainty in optimal actions.

Considering the observations above, it is obvious that PlaNet is insufficiently uncertainty-aware, which leads to limitations on its strong potential.
Although the training procedure employed in PlaNet is based on autoencoding variational inference~\cite{kingma2013auto}, the model parameters are estimated as fixed-points $\theta$, underestimating the model uncertainty $p(\theta|\mathcal{D})$.
For the purpose of planning, the cross entropy method (CEM)~\cite{botev2013cross} is heuristically introduced for trajectory optimization, which ignores the multimodal uncertainty in optimal actions.

Motivated by this, in the present study, we reconsider PlaNet from a Bayesian viewpoint, aiming to propose an extension of PlaNet, referred to as \textit{PlaNet of the Bayesians} (PlaNet-Bayes).
The primary contributions, which characterize PlaNet-Bayes, can be summarized as follows;
\begin{itemize}
  \item We propose incorporating uncertainty in latent state-space models by considering approximate inference of the posterior $p(\theta|\mathcal{D})$ with a neural network ensemble.
  \item We formulate VI-MPC for partially observable environments by considering latent planning as variational inference.
  Exploiting the newly derived framework, we also introduce a latent version of PaETS to involve multimodality in optimal actions.
\end{itemize}
The differences between the proposed method and the baseline PlaNet are summarized in the bottom part of Fig.~\ref{fig:fig1}.
By involving the two types of the uncertainty, PlaNet-Bayes can ahieve the performance consistently better compared with PlaNet. 

The remainder of this paper is organized as follows.
In Sec.~\ref{sec:preliminary}, we provide a brief review on PlaNet, VI-MPC, and PaETS.
In Sec.~\ref{sec:proposal}, we describe the proposed PlaNet-Bayes in detail.
In Sec.~\ref{sec:experiments}, the effectiveness of PlaNet-Bayes is demonstrated through evaluations using the DeepMind control suite~\cite{deepmindcontrolsuite2018}.


\section{Preliminary} \label{sec:preliminary}
\newcommand{\pjoint}{p_{\operatorname{joint}}}
\newcommand{\predtraj}{p(\latent_{\leq t+T}|\action_{\leq t+T-1})}
\newcommand{\predtrajt}{p(\latent|\action,\theta)}
\newcommand{\predtrajs}{p(\latent_{\leq t}|\action_{< t})}
\newcommand{\predtrajst}{p(\latent_{\leq t}|\action_{< t}, \theta)}
\newcommand{\likelihood}{p(\obs_{\leq t} | \latent_{\leq t})}
\newcommand{\actiongt}{\action_{\geq t}}
\newcommand{\actionlt}{\action_{< t}}
\newcommand{\encoderm}{q(\latent_{\leq t}|\obs_{\leq t}, \action_{\leq t})}
\newcommand{\encodermt}{q(\latent_{\leq t}|\obs_{\leq t}, \action_{\leq t}, \theta)}

\subsection{PlaNet: Deep Planning Network}
\subsubsection{Autoencoding Variational Bayes for Time Series}
\begin{figure}[tb]
  \centering
  \includegraphics[width=0.4\textwidth]{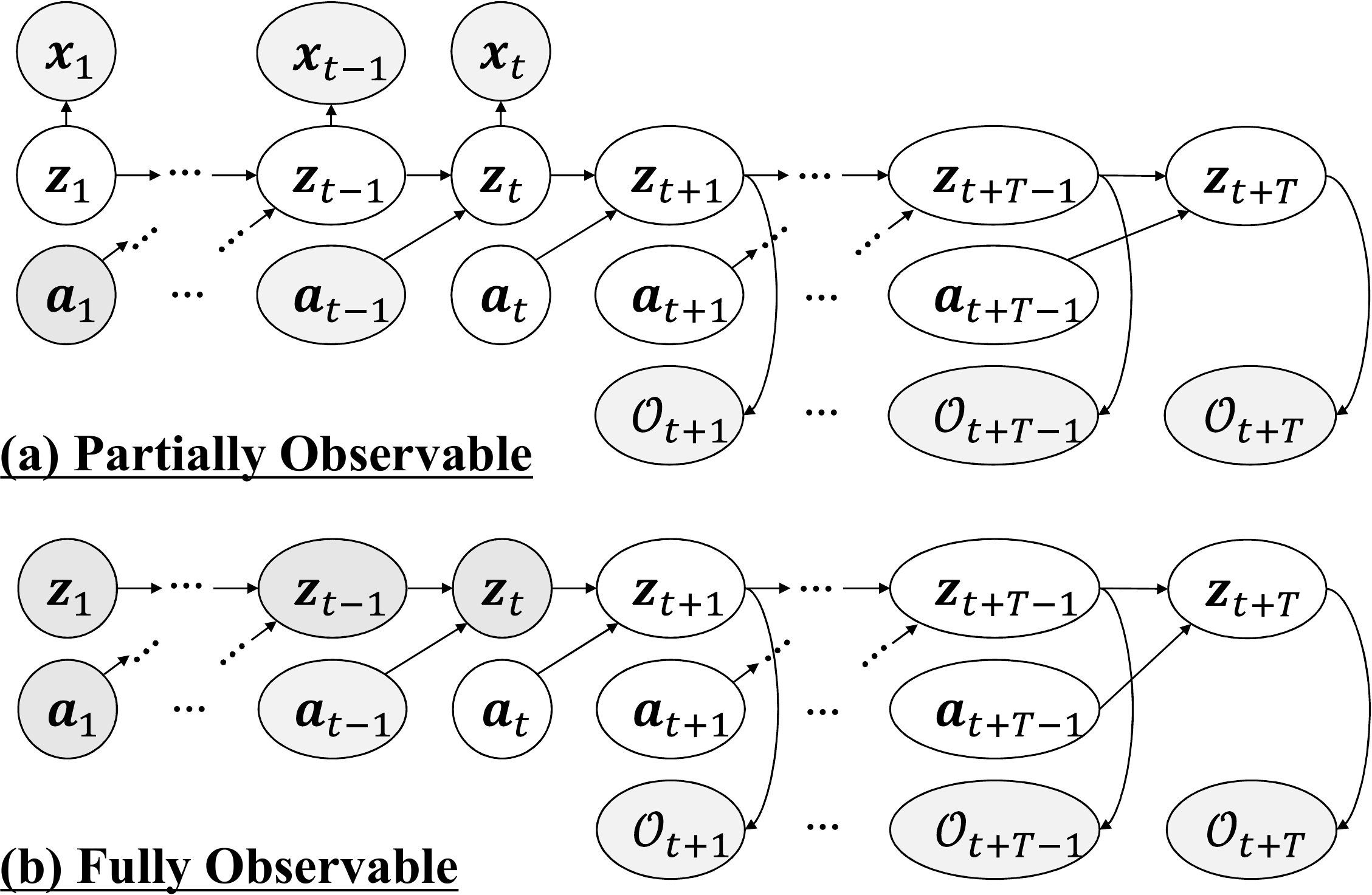}
  \caption{Graphical models discussed in this paper where $t$ is the current time-step and $T$ is the planning horizon. The training objective of PlaNet and VI-MPC for partially observable environments are derived from (a). VI-MPC for fully observable environments is derived from (b).}
  \label{fig:graphical_model}
\end{figure}
Let us begin with considering the graphical model illustrated in Fig.~\ref{fig:graphical_model}(a).
In this section, we focus on $\latent_{\leq t}$ and their adjacent nodes.
The remaining nodes, namely $\latent_{>t}$, $\action_{\geq t}$, and $\mathcal{O}_{> t}$, are discussed later in Sec.~\ref{sec:latent_planning}.
The joint distribution of the focused variables is defined as follows:
\begin{align}
  & \pjoint(\latent_{\leq t}, \action_{<t}, \obs_{\leq t}) = \nonumber\\
  & \underbrace{
    p(\latent_{1})
    \left\{\prod_{t'=1}^{t-1} p\left(\latent_{t'+1} | \latent_{t'}, \action_{t'} \right)\right\}
  }_{\coloneqq \predtrajs}
    \cdot
    \underbrace{
      \left\{\prod_{t'=1}^{t} p\left(\obs_{t'} | \latent_{t'} \right)\right\}.
    }_{\coloneqq \likelihood}
\end{align}
As in the case of well-known variational autoencoders (VAEs)~\cite{kingma2013auto}, generative models $p(\latent_{t+1}|\latent_{t}, \action_{t})$, $p(\obs_{t}|\latent_{t})$ and inference model $q(\latent_{t}|\obs_{\leq t}, \action_{< t})$ can be trained by maximizing the
evidence lower bound (ELBO):
\begin{align}
  & \log p(\obs_{\leq t} | \action_{\leq t}) = \log \int \predtrajs \likelihood d\latent_{\leq t} \nonumber\\
  \geq & \underbrace{\mathbb{E}_{\encoderm}\left[\log \likelihood \right]}_{\mathrm{reconstruction}} \hookleftarrow\nonumber \\
  & - \underbrace{D_{\operatorname{KL}}\left[\encoderm || \predtrajs\right]}_{\mathrm{complexity}}, \label{eqn:vae_elbo}
\end{align}
where,
\begin{equation}
  \encoderm \coloneqq \prod^{t}_{t'=1} q(\latent_{t'}| \obs_{\leq t'}, \action_{< t'}). \label{eqn:inference}
\end{equation}
If the models are parameterized with $\theta$, this objective can be maximized by the stochastic gradient ascent via backpropagation.
For the purpose of latent planning, a reward function $p(r_{t}|\latent_{t})$ is also required to be formulated.
To do this, we can simply regard the rewards as observations and learn the reward function along with $p(\obs_{\latent}|\latent_{t})$.

\subsubsection{Recurrent State-Space Model}
\begin{figure}[tb]
  \centering
  \includegraphics[width=0.45\textwidth]{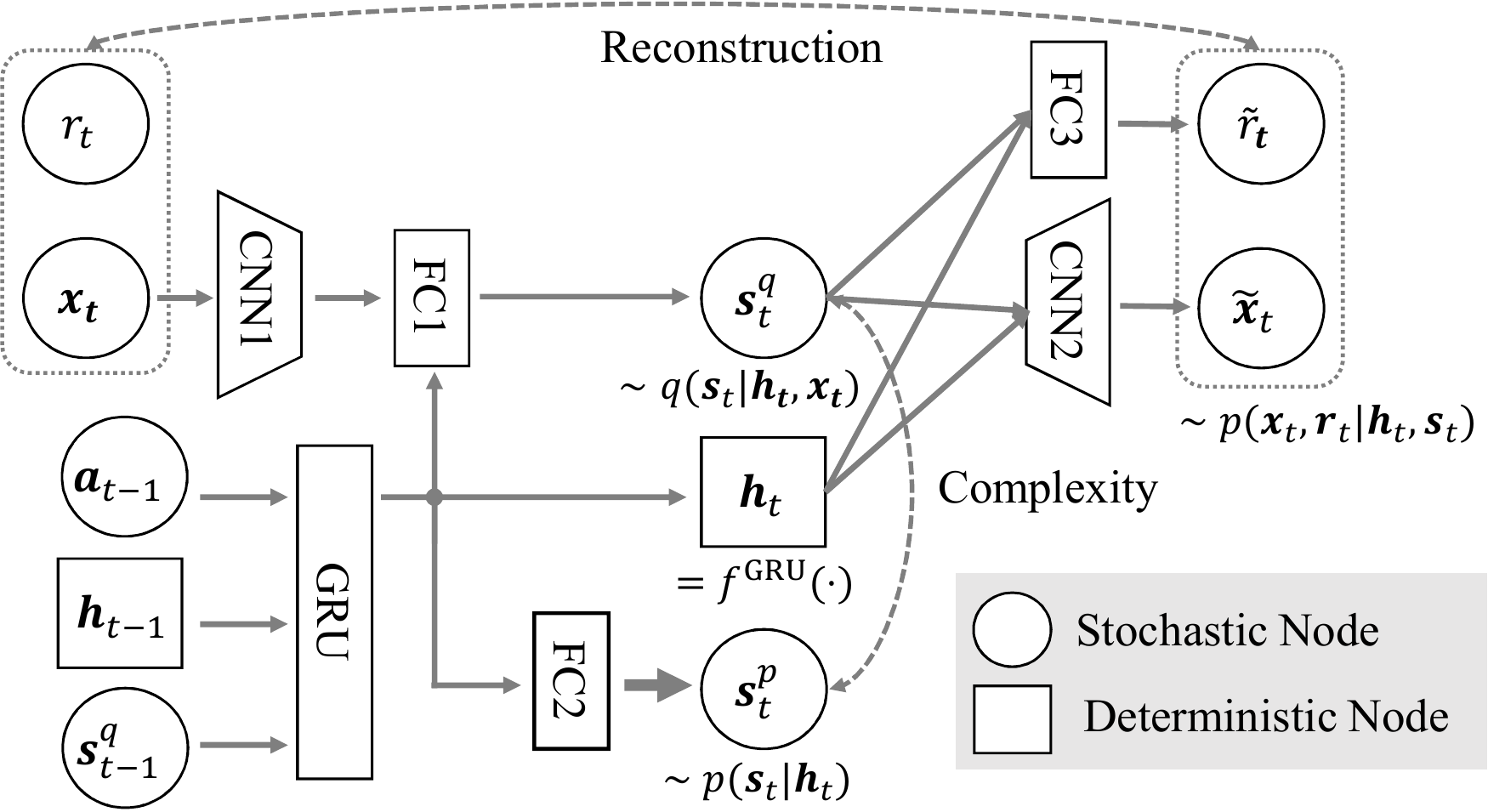}
  \caption{The architecture of RSSM. \texttt{CNN}, \texttt{GRU} and \texttt{FC} represent a convolutional neural network, a GRU-cell, and a fully-connected layer, respectively.
}
  \label{fig:training_scheme}
\end{figure}

PlaNet introduces Recurrent State-Space Model (RSSM), which assumes the latent $\latent_{t}$ which comprises $\latent_{t} = (\latents_{t} \latenth_{t})$
where $\latents_{t}$, $\latenth_{t}$ are the probabilistic and deterministic variable, respectively.
The generative and inference models can be formulated as:
\begin{align}
  \mathrm{Generative\ models}&:
  \begin{cases}
    \latenth_{t} = f^{\mathrm{GRU}}(\latenth_{t-1}, \latents_{t-1}, \action_{t-1}) \\
    \latents_{t} \sim p(\latents_{t} | \latenth_{t}) \\
    \obs_{t}, r_{t} \sim p(\obs_{t}, r_{t} | \latenth_{t}, \latents_{t})
  \end{cases}, \nonumber\\
  \mathrm{Inference\ model}&: \latents_{t} \sim q(\latents_{t} | \latenth_{t}, \obs_{t}),
\end{align}
where deterministic $\latenth_{t}$ is considered as the internal state of the gated recurrent unit (GRU) $f^{\mathrm{GRU}}(\cdot)$~\cite{cho2014learning}, so that historical information is embedded into $\latenth_{t}$.
The architectures of the generative and inference models are illustrated in Fig.~\ref{fig:training_scheme}.

\subsubsection{Latent Planning using RSSM}
Algorithm~\ref{alg:planet_cem} outlines PlaNet's latent planning strategy with CEM~\cite{botev2013cross}.
CEM is a stochastic method based on importance sampling that is used to iteratively update a proposal distribution to optimize an objective function by executing the following steps:
(1) sample $K$ candidates from the proposal distribution,
(2) then evaluate the objective function for each sample, and
(3) update the proposal distribution using the results of evaluation.
Algorithm~\ref{alg:planet_cem} summarizes the process executed by PlaNet to implements the above steps for latent planning.
(1) At $\ell$5, $K$ candidate action sequences are sampled from the diagonal Gaussian proposal parameterized with mean $\actionm_{t:t+T-1}$ and variance $\actions_{t:t+T-1}$.
In the pseudo-code, the subscript $\Box_{t:t+T-1}$ is omitted for readability.
(2) Starting from the current state $(\latenth_{t}, \latents_{t})$ estimated at $\ell$1--3, latent trajectories and step rewards are sampled up to $T$ time-steps at $\ell$6--10.
(3) Using the trajectory rewards calculated at $\ell$11,
the parameters $\actionm$, $\actions$ are adjusted according to the CEM update law at $\ell$12--14, where $\mathds{1}[\cdot]$ is an indicator function and $R_{\mathrm{thd}}$ is determined so that top-$e\%$ of samples satisfy the threshold condition ($e=10\%$ is used in \cite{hafner2018learning}).
This iterative update is executed $U$ times.
\SetAlCapNameFnt{\footnotesize}
\SetAlCapFnt{\footnotesize}

\begin{algorithm}[h]
\footnotesize
\DontPrintSemicolon
  \KwInput{Current observation $\obs_{t}$, \\
  Previous latent states and action $(\latenth_{t-1}, \latents_{t-1}, \action_{t-1})$, \\
  Initial parameters of the proposal distribution~$(\actionm^{(0)}, \actions^{(0)})$}
  \KwOutput{Optimized parameters of proposal dist.~$(\actionm^{(U)}, \actions^{(U)})$ \\
  Current latent states $(\latenth_{t}, \latents_{t})$}
  \tcp*[l]{(0) Estimate current latent state}
  \If{$t > 1$}{
  Evolve the latent state $\latenth_{t} = f^{\mathrm{GRU}}(\latenth_{t-1}, \latents_{t-1}, \action_{t-1})$ \;
  }
  Infer the latent state $\latents_{t} \sim q(\latents_{t} | \latenth_{t}, \obs_{t})$ \;
  \For{$j \leftarrow 1$ \KwTo $U$}{
  \tcp*[l]{(1) Sample $K$ candidates}
  Sample actions $\{\action_{k} \sim q(\action; \actionm^{(j-1)}, \actions^{(j-1)}\}_{k=1}^{K}$ \;
  \tcp*[l]{(2) Eval.~the objective for each candidate}
  Sample trajectories and rewards $\{ \{$ \;
  \Indp $\latenth_{k, t'+1} = f^{\mathrm{GRU}}(\latenth_{t'}, \latents_{t'}, \action_{k,t'})$, \;
  $\latents_{k,t'+1} \sim p(\latents_{t'+1} | \latenth_{k, t'+1})$ \;
  $r_{k,t'+1} \sim p(r_{t'+1} | \latenth_{k, t'+1}, \latents_{k, t'+1})$ \;
  \Indm $\}_{t'=t}^{t+T-1} \}_{k=1}^{K}$ \;
  Calc.~trajectory rewards $\{R_{k} = \sum^{t+T}_{t'=t+1}r_{k,t'}\}^{K}_{k=1}$\;
  \tcp*[l]{(3) Update proposal distribution}
  Calc.~weights $\{w_{k} = \mathds{1}[R_{k} \geq R_{\mathrm{thd}}] \}^{K}_{k=1}$ \;
  Normalize weights $\{w_{k} \leftarrow w_{k} / \sum^{K}_{k'=1}w_{k'}\}^{K}_{k=1}$ \;
  Update $(\actionm^{(j)}, (\actions^{(j)})^{2}) \leftarrow (\sum^{K}_{k=1}w_{k}\cdot\action_{k}, \sum^{K}_{k=1}w_{k}\cdot(\action_{k}-\actionm^{(j)})^{2})$ \;
  }
\caption{PlaNet's latent planning based on CEM \cite{hafner2018learning}} \label{alg:planet_cem}
\end{algorithm}

\subsection{VI-MPC and PaETS in Fully Observable Environments} \label{sec:vimpc}
%
%
Figure \ref{fig:graphical_model}(b) represents the graphical model that is used to derive original VI-MPC~\cite{okada2019variational} based on the \textit{control as inference} framework~\cite{levine2018reinforcement}.
To formulate optimal control as inference, a binary random variable $\mathcal{O}_{t'} \in \{0, 1\}$ is auxiliarily introduced
to represent the \textit{optimality} of state $\latent_{t'}$. Note that $\latent_{t'}$ is not \textit{latent} state here.
Let us consider the trajectory posterior conditioned on the optimality $p(\action_{\geq t}, \latent_{>t} | \mathcal{O}_{>t})$.
Hereinafter, we denote $\action_{\geq t}$, $\latent_{> t}$ and $\mathcal{O}_{> t}$ as; $\action$, $\latent$ and $\mathcal{O}$ for readability.
By solving a variational inference problem: $\operatorname{argmin}_{q} \operatorname{KL}(q(\action, \latent) || p(\action, \latent | \mathcal{O}) )$, we obtain the iterative law to update the variational distribution $q$ as per equation bellow:
\begin{equation}
  q^{(j+1)}(\action) \leftarrow
  \frac
  {q^{(j)}(\action)\cdot \mathcal{W}(\action)^{\frac{1}{\lambda}}\cdot (q^{(j)}(\action))^{-\kappa}}
  {\mathbb{E}_{q^{(j)}(\action)}\left[\mathcal{W}(\action)^{\frac{1}{\lambda}}\cdot (q^{(j)}(\action))^{-\kappa}\right]}, \label{eqn:vimpc}
\end{equation}
where $j$ indicates the loop count, $\lambda$ is the inverted step-size to control optimization speed, and $\kappa$ is the weight of the entropy regularization term $q^{-\kappa}$.
$\mathcal{W}$ is defined as:
\begin{equation}
  \mathcal{W}(\action) \coloneqq  \mathbb{E}_{p(\latent|\action)}[
  p(\optimal | \latent) \label{eqn:weight0}
  ].
\end{equation}
This is a general framework called VI-MPC that generalizes several MPC methods.
Different definitions of optimality likelihood $p(\mathcal{O} | \latent)$ recover various methods including
CEM, path integral control \cite{okada2017path,williams2016aggressive},
covariance  matrix  adaptation  evolution  strategy  (CMA-ES)~\cite{hansen2003reducing},
and proportional CEM~\cite{goschin2013cross}.
The above-mentioned methods generally assume that $q$ is Guassian.
However, VI-MPC can define $q$ arbitrarily, and PaETS defines $q$ as GMM successfully incorporating multimodality of optimal actions.
The original formulation of VI-MPC supposes fully observable environments, and application to partially observable cases has not been discussed.

\section{PlaNet-Bayes: PlaNet of the Bayesians} \label{sec:proposal}
This section describes an extension of PlaNet referred to as \textit{PlaNet of the Bayesians} (PlaNet-Bayes) that implies incorporating two types of uncertainty in latent models and actions by exploiting Bayesian inference.
The process of incorporating both types of uncertainty is introduced in Sections~\ref{sec:model_uncertainty} and \ref{sec:action_uncertainty}.

\subsection{Incorporating Model Uncertainty} \label{sec:model_uncertainty}
In general, VAEs denote generative and inference models like $p_{\theta}(\cdot)$, $q_{\theta}(\cdot)$,
assuming a point estimation of the model parameters $\theta$.
Here, we remove this assumption and treat $\theta$ as a random variable.
As indicated in Eq.~\ref{eqn:vae_elbo}, maximizing ELBO also leads to maximization of $p(\mathcal{D}|\theta)$,
and therefore we can regard VAEs' general training procedure as a maximum likelihood estimation.
Instead, by inferring the posterior $\modelpost$, we incorporate the uncertainty in the model parameters of RSSM.
Given a sufficiently parameterized model, i.e., a deep neural network, promising schemes for approximating the posterior are
stochastic gradient MCMC~\cite{daxberger2019bayesian},
\textit{dropout as variational inference} \cite{gal2016dropout,gal2017concrete,kahn2017uncertainty},
and neural network ensembles~\cite{chua2018deep,kurutach2018model,clavera2018model,okada2019variational}.

In this paper, we use the ensemble scheme
owing to its simplicity and better performance compared with dropout~\cite{okada2019variational}.
This scheme is employed to approximate the posterior as a set of \textit{particles} $\modelpost \simeq \frac{1}{E}\sum^{E}_{i}\delta(\theta - \theta_{i})$,
where $E$ is the ensemble size (namely, the number of networks) and $\delta$ is Dirac delta function.
This approximation can successfully incorporate multimodal uncertainty in the exact posterior.
Each particle $\theta_{i}$ is independently trained by stochastic gradient descent so as to (sub-)optimize
$\modelpost \propto p(\mathcal{D}|\theta)p(\theta)$.
%
%
In the proposed modeling scheme, \texttt{GRU}, \texttt{FC1}, and \texttt{FC2}, as presented in Fig.~\ref{fig:training_scheme}, are implemented as ensemble networks, and the ensemble size is set to be $E=5$ as same as defined in~\cite{okada2019variational,chua2018deep}.

\subsection{Incorporating Action Uncertainty} \label{sec:action_uncertainty}
\subsubsection{Derivation of latent VI-MPC} \label{sec:latent_planning}
%
%
We derive VI-MPC for partially observable environments by formulating latent planning as a variational inference problem.
We refer to the graphical model of Fig.~\ref{fig:graphical_model}(a) again for this formulation.
For the purpose of clarity, in this figure, we omit the random variable $\theta$ using which the latent state transition is conditioned,
namely, $p(\latent_{t}|\latent_{t-1}, \action_{t-1}, \theta)$.
The joint distribution of all variables in this figure is:
\begin{align}
    & \pjoint(\latent, \action, \theta, \optimal_{\geq t}, \obs_{\leq t}) = \nonumber\\
    & {p(\optimal_{> t}| \latent_{> t})}
    \cdot
    {\predtrajt}
    \cdot
    {\likelihood}
    \cdot
    \modelpost
  \label{eqn:traj_posterior},
\end{align}
where,
\begin{equation}
  p(\optimal_{> t}| \latent_{> t}) = \prod^{t+T}_{t'=t+1} p(\optimal_{t'}| \latent_{t'}),
\end{equation}
and a non-informative action prior $p(\action)$ is supposed.
The subscripts of $\latent_{1:t+T}$, $\action_{1:t+T-1}$ are omitted.
The objective of this formulation is to infer the posterior:
\begin{equation}
  p(\action_{\geq t}, \latent, \theta | \optimal_{\geq t}, \obs_{\leq t}) =
  \frac
  {\pjoint(\cdot)}
  {\int\pjoint(\cdot)d\action_{\geq t}d\latent d\theta}.
\end{equation}
As the inference of this posterior is intractable, instead, we estimate the variational distribution $q(\action_{\geq t}, \latent, \theta)$ that
minimizes $\operatorname{KL}(q(\action_{\geq t}, \latent, \theta) || p(\action_{\geq t}, \latent, \theta | \optimal_{> t}, \obs_{\leq t}))$.
We suppose that $q$ is factorized as:
\begin{align}
  & q(\action_{\geq t}, \latent, \theta) = \nonumber \\
  & q(\action_{\geq t})  \predtrajt \likelihood \modelpost. \label{eqn:variational_dist}
\end{align}
This variational inference problem can be solved by maximizing the ELBO:
\begin{align}
  \log p(\optimal_{> t}, \obs_{\leq t} ) \geq \mathbb{E}_{q(\actiongt), \predtrajt, \modelpost}[ \hookleftarrow \nonumber\\
  \likelihood p(\optimal_{\geq t}| \latent_{\geq t}) - q(\actiongt)]. \label{eqn:elbo_vimpc}
\end{align}
By applying the mirror descent~\cite{bubeck2015convex,okada2018acceleration,okada2019variational} to this optimization problem, we can derive an update law for $q(\actiongt)$,
which takes the same form as Eq.~\ref{eqn:vimpc}.
At the same time, $\mathcal{W}(\actiongt)$ is defined differently from the original formulation:
%
\begin{equation}
  \mathcal{W}(\actiongt) \coloneqq  \mathbb{E}_{\predtrajt, \modelpost}[ 
  \likelihood p(\optimal_{> t}| \latent_{> t}) \label{eqn:weight}
  ].
\end{equation}
A major difference from the original VI-MPC is that
the observation likelihood $\likelihood$ should be considered in Eq.~\ref{eqn:weight}.
One may consider that Eq.~\ref{eqn:weight} can be efficiently implemented with Sequential Monte Carlo (SMC also referred to as particle filter)~\cite{thrun2002probabilistic}
using the generative models; namely, for $t'\leq t$,
predict particles by $p(\latent_{t'}|\latent_{t'-1}, \action_{t'-1}, \theta)$ and conduct resampling with the likelihood $p(\obs_{t'}|\latent_{t'})$.
Because of the fact that this yields rather a complicated process, let us consider to approximate it by utilizing the inference model.
By applying importance sampling with $\encodermt$, 
Eq.~\ref{eqn:weight} can be rearranged as:
\begin{align}
  & \mathcal{W}(\actiongt) = \nonumber \\
  & \mathbb{E}_{\mathbb{P}}\left[
  \frac{\predtrajst}{\encodermt} \likelihood p(\optimal_{\geq t}| \latent_{\geq t}) \label{eqn:weight2}
  \right],
\end{align}
where,
\begin{equation}
  \mathbb{P} \coloneqq
  \encodermt p(\latent_{> t} | \action_{\geq t}, \theta) \modelpost. \label{eqn:latent_sampler}
\end{equation}
Having this distribution $\mathbb{P}$, $z_{\leq t}$ and $z_{> t}$ are sampled from
the inference model $q(\latent_{t}|\cdot)$ and generative model $p(\latent_{t}|\cdot)$, respectively.
As these models are trained so as to minimize the complexity loss in Eq.~\ref{eqn:vae_elbo},
we can expect that the likelihood ratio in Eq.~\ref{eqn:weight2}, i.e., $p(\latent_{\leq t}|\cdot) / q(\latent_{\leq t}|\cdot)$, can be canceled.
In addition, minimization of the reconstruction loss in Eq.~\ref{eqn:vae_elbo} allow realizing successful autoencoding $\obs\rightarrow \latent \rightarrow \obs$ so that
\begin{equation}
  \likelihood \simeq \textrm{const.}\ \textrm{if}\ \latent_{\leq t} \sim \encodermt
\end{equation}
Therefore, the likelihood $\likelihood$ can be approximately pulled out from $\mathbb{E}_{\mathbb{P}}[\cdot]$ and considered as canceled in Eq.~\ref{eqn:vimpc}.
On the basis of these observations, we rewrite Eq.~\ref{eqn:weight2} as:
\begin{equation}
  \mathcal{W}(\actiongt) \coloneqq \mathbb{E}_{\mathbb{P}}\left[p(\optimal_{\geq t}| \traj_{\geq t})\right].
  \label{eqn:weight3}
\end{equation}
Eqs.~\ref{eqn:vimpc}, \ref{eqn:latent_sampler}, and \ref{eqn:weight3} suggest general VI-MPC for the latent state-space models.
The planning method used in PlaNet is a special case of VI-MPC, assuming that
(1) the variational distribution $q$ is Gaussian;
(2) the optimality likelihood is defined as: $p(\optimal_{\geq t}| \latent_{\geq t}) \coloneqq \mathds{1}[R(\latent_{\geq t}) \geq R_{\mathrm{thd}}]$ ($R(\latent_{\geq t})$: trajectory reward);
(3) $\mathbb{E}_{\mathbb{P}}[\cdot]$ is approximated by a single sample;
(4) $\modelpost$ is inferred as a single point; and (5) the entropy regularizer $\kappa$ is set to be $\kappa \rightarrow 0$.
The generality of original VI-MPC holds in this partially observable case,
and we can utilize different definitions of $q$ and $p(\optimal_{\geq t}| \latent_{\geq t})$.
In this study, same optimality likelihood with PlaNet is introduced.

\subsubsection{PaETS for latent planning}
We introduce a latent version of PaETS using the proposed latent VI-MPC.
PaETS uses GMM as the variational distribution $q$ as:
\begin{equation}
q^{(j)}(\actiongt) \coloneqq q(\actiongt; \phi^{(j)}) = \sum^{M}_{m=1}\pi^{(j)}_{m} \mathcal{N}(\actiongt; \actionm^{(j)}_{m}, \actions^{(j)}_{m}), \label{eqn:q_gmm}
\end{equation}
where
$\phi^{(j)} \coloneqq \{(\mixcoef_{m}^{(j)}, \actionm_{m}^{(j)}, \actions_{m}^{(j)} )\}^{M}_{m=1}$ and $M$ is the number of components in the mixture model.
By following the derivation procedure similarly as in \cite{okada2019variational},
we obtain the update laws of $\phi^{(j+1)}$, which take the weight-average form like $\ell 14$ of Algorithm~\ref{alg:planet_cem}:
\begin{align}
\textstyle
& \left(\actionm_{m}^{(j+1)}, \actions_{m}^{(j+1)}, \mixcoef_{m}^{(j+1)} \right)
\leftarrow \nonumber\\
& \left(
\textstyle
\sum_{k} \omega^{(j+1)}_{m,k}\action_{k},
{\sum_{k} \omega^{(j+1)}_{m,k}(\action_{k} - \actionm_{m}^{(j+1)})^{2}},
\frac{N_{m}}{\sum^{M}_{m'=1} N_{m'}}
\right). \label{eqn:paets}
%
%
\end{align}
The complete definition of Eq.~\ref{eqn:paets} is available in  Appendix~\ref{sec:paets_eqs}.
\subsection{Procedure and Implementation} \label{sec:procedure}
Algorithms~\ref{alg:bplanet_outer} and \ref{alg:bplanet_inner} summarize the steps of the proposed method.
The outermost loop of Algorithm~\ref{alg:bplanet_outer} describes the training of the posterior $\modelpost$ (namely, neural network ensemble), at which the model is iteratively trained at $\ell$3. The trained model is tested on the MPC loop at $\ell$6--11, whereas $\mathcal{D}$ is augmented according to the executed trajectories.
This paper realizes GMM parameter initialization at $\ell 7$ as follows;
$\actionm^{(0)}_{m}$ is reset by a general warm-start technique of MPC,
and $(\actions^{(0)}_{m}, \mixcoef^{(0)}_{m})$ is set to be $(\mathbf{1}/2, 1/M)$.

The procedure in Algorithm~\ref{alg:bplanet_inner} is rather similar to that one of in Algorithm~\ref{alg:planet_cem}. One of the primary differences is that the expectation $\mathbb{E}_{\mathbb{P}}[\cdot]$ in Eq.~\ref{eqn:weight3} is approximated with $E$ samples obtained from the $E$ ensemble networks, whereas PlaNet uses only a single sample.
Another difference is that PaETS is employed to update the variational distribution parameter $\phi$ at $\ell$12%
\footnote{
It should be noted that the heuristics discussed in \cite{okada2019variational} is introduced at $\ell$11--12.
Let us describe the optimality likelihood as $p(\optimal_{\geq t}| \latent_{\geq t}) \coloneqq f^{\mathrm{opt}}(R(\latent))$,
where $f^{\mathrm{opt}}(\cdot)$ is a monotonic increasing function and $R(\latent)$ is a trajectory reward function.
The heuristics use $\mathcal{W}'(\actiongt) \coloneqq f^{\mathrm{opt}}(\mathbb{E}[R(\latent)])$ instead of $\mathcal{W}(\actiongt)\coloneqq \mathbb{E}[f^{\mathrm{opt}}(R(\latent))]$.
It has been experimentally observed that these heuristics demonstrates higher optimization performance~\cite{okada2019variational,chua2018deep}.
}.

\begin{algorithm}[h]
\footnotesize
\DontPrintSemicolon
  Initialize $\mathcal{D}$ with a random controller for several trials \;
  \tcp*[l]{Training loop}
  \Repeat{the MPC-policy performs well}{
  Train $\{\theta_{i}\}^{E}_{i=1}$ by optimizing Eq.~\ref{eqn:vae_elbo} \; 
  Initialize $\{(\latenth_{i,0}, \latents_{i,0})\}^{E}_{i=1}$ and $\action_{0}$\ as $\mathbf{0}$s \;
  Reset environment and observe $\obs_{1}$ \;
  \tcp*[l]{Control loop, $H$: episode length}
  \For{$t \leftarrow 1$ \KwTo $H$}{
  Initialize GMM parameter $\phi^{(0)}$ \;
  Execute Alg.~\ref{alg:bplanet_inner} \;
  Sample $\action_{t:t+T-1} \sim q(\action_{t:t+T-1}; \phi^{(U)})$ \;
  Send $\action_{t}$ to actuators and observe $(\obs_{t+1}, r_{t+1})$\;
  $\mathcal{D} \leftarrow \mathcal{D} \cup \{(\obs_{t}, \action_{t}, r_{t+1})\}$ \;
  }}
\caption{PlaNet-Bayes' training and control loop} \label{alg:bplanet_outer}
\end{algorithm}
\begin{algorithm}[h]
\footnotesize
\DontPrintSemicolon
  \KwInput{Current observation $\obs_{t}$, \\
  Previous latent states and action $\{(\latenth_{i, t-1}, \latents_{i, t-1})\}^{E}_{i=1}$, $\action_{t-1}$, \\
  Initial parameters of variational distribution~$\phi^{(0)}$}
  \KwOutput{Optimized parameters of variational distribution~$\phi^{(U)}$ \\
  Current latent states $\{(\latenth_{i, t}, \latents_{i, t})\}^{E}_{i=1}$}
  \tcp*[l]{(0) Estimate the current latent state}
  \If{$t > 1$}{
  Evolve the latent state $\{\latenth_{i,t} = f^{\mathrm{GRU}}(\latenth_{i,t-1}, \latents_{i,t-1}, \action_{t-1}; \theta_{i})\}^{E}_{i=1}$ \;
  }
  Infer the latent state $\{\latents_{i,t} \sim q(\latents_{t} | \latenth_{i,t}, \obs_{t}, \theta_{i})\}^{E}_{i=1}$ \;
  \For{$j \leftarrow 1$ \KwTo $U$}{
  \tcp*[l]{(1) Sample $K$ candidates}
  Sample actions $\{\action_{k} \sim q(\action; \phi^{(j-1)}\}_{k=1}^{K}$ \;
  \tcp*[l]{(2) Eval.~the objective for each candidate}
  Sample trajectories and rewards $\{ \{ \{$ \;
  \Indp $\latenth_{k, i, t'+1} = f_{\theta_{i}}^{\mathrm{GRU}}(\latenth_{k,i,t'}, \latents_{k,i,t'}, \action_{k,t'})$, \;
  $\latents_{k,i,t'+1} \sim p(\latents_{k,t'+1} | \latenth_{k, i, t'+1}, \theta_{i})$ \;
  $r_{k,i,t'+1} \sim p(r_{t'+1} | \latenth_{k, i, t'+1}, \latents_{k, i, t'+1})$ \;
  \Indm $\}_{t'=t}^{t+T-1} \}_{i=1}^{E}\}_{k=1}^{K}$ \;
  Calc.~trajectory rewards $\{R_{k} = \frac{1}{E}\sum^{E}_{i=1}\sum^{t+T}_{t'=t+1}r_{k,i,t'}\}^{K}_{k=1}$\;
  \tcp*[l]{(3) Update proposal distribution}
  Update $\phi^{(j)}$ by Eq.~\ref{eqn:paets}.
  }
\caption{PlaNet-Bayes' latent planning with PaETS} \label{alg:bplanet_inner}
\end{algorithm}

We implement PlaNet-Bayes in TensorFlow~\cite{tensorflow2015-whitepaper} by modifing the official source code of PlaNet%
\footnote{\url{https://github.com/google-research/planet}}.
We keep the most of hyperparameters and experimental conditions similar as the original ones except for the number of action candidates $K$ (originally $K=1,000$).
As $\mathbb{E}_{\mathbb{P}}[\cdot]$ is approximated with $E=5$ samples, we have to sample $K\times E$ trajectories at $\ell$6--10 in Algorithm~\ref{alg:bplanet_inner}, which requires $E$-times larger computations.
To relieve this, we set $K \leftarrow K / E (=1,000/5=200)$ so that the same number of trajectories are evaluated; namely the total number of multiply-accumulate operations are kept same.
The other hyperparameters introduced in this papers are $\kappa=0$, $\lambda=1$, and $M=5$.

\section{Experiments} \label{sec:experiments}
\subsection{Comparison with PlaNet}
\begin{figure}[tb]
  \centering
  \includegraphics[width=0.275\textwidth]{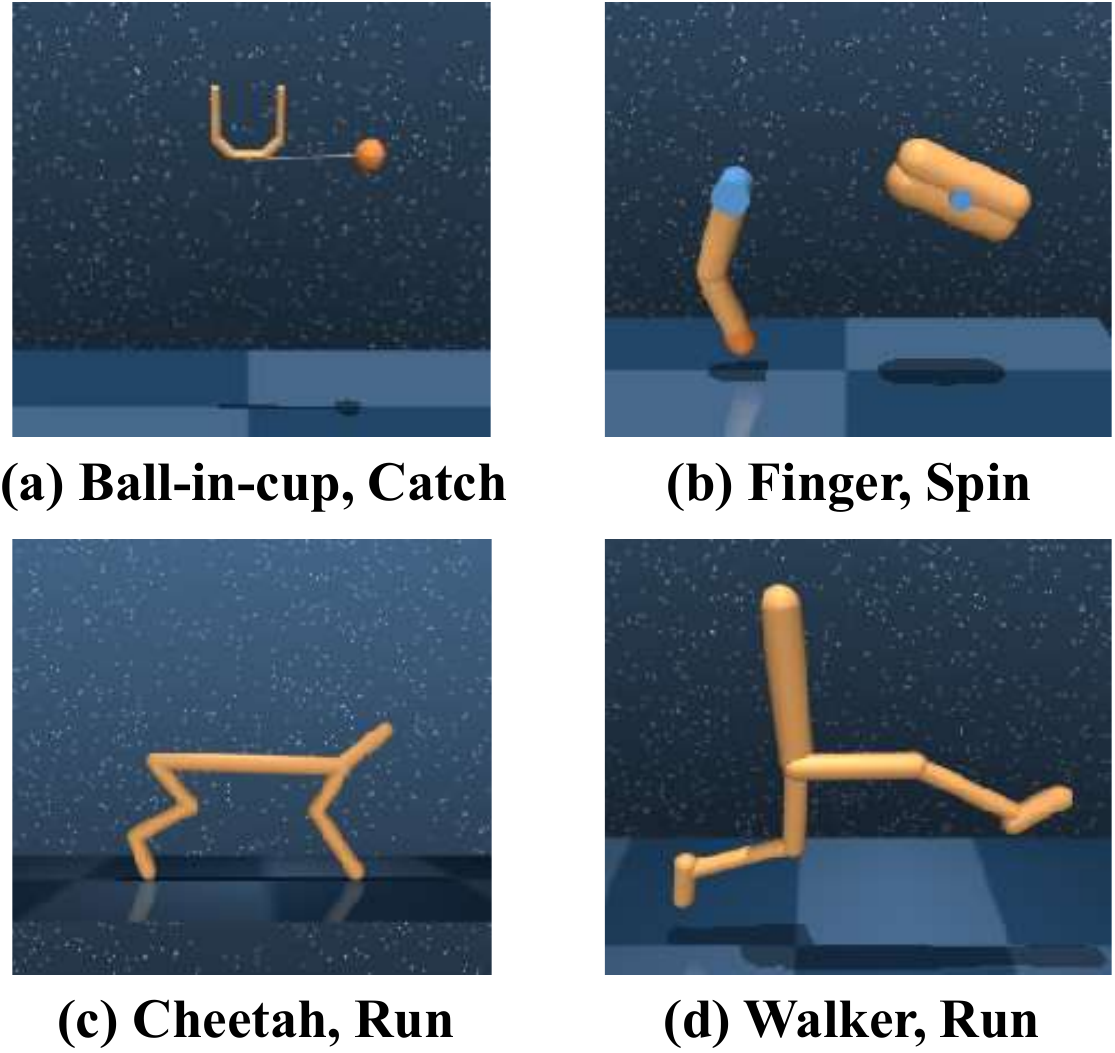}
  \caption{Four experimental control tasks from the DeepMind Control Suite~\cite{deepmindcontrolsuite2018}.
  In all tasks, only third-person views (i.e., $64\times 64\times 3$ pixels) are input into the MPC-policy as observations $\obs$.
  }
  \label{fig:tasks}
\end{figure}
\begin{figure}[t]
  \centering
  \includegraphics[width=0.475\textwidth]{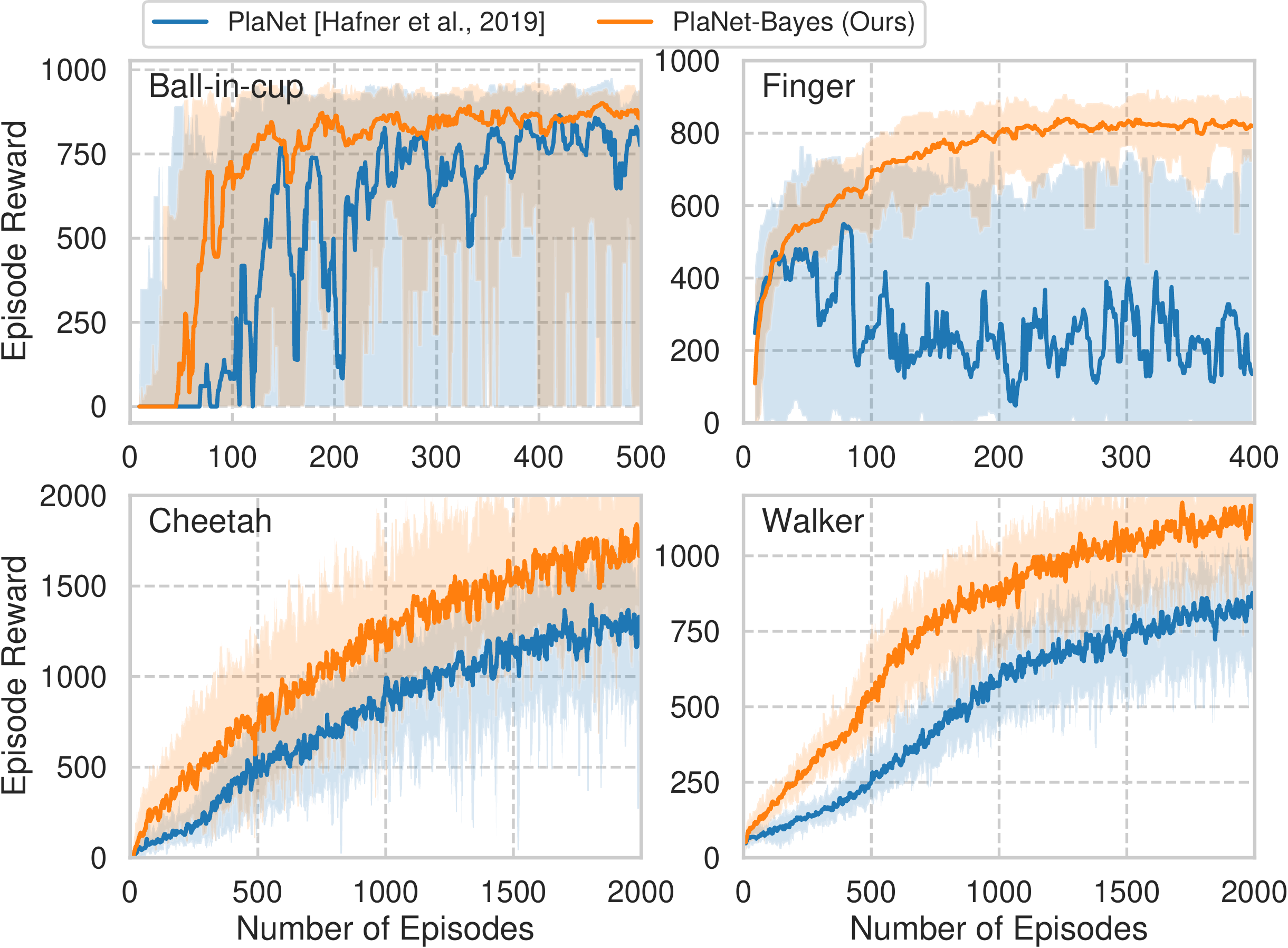}
  \caption{
  Comparison between PlaNet-Bayes and the baseline PlaNet.
  The lines represent the medians, and the shaded areas depict the percentiles 5 to 95 over 4 seeds and 10 trajectories.
  The discrepancy in the performance of PlaNet with respect to \cite{hafner2018learning} is caused by our task modification$^{\ref{footnote:task}}$.
  }
  \label{fig:comp_sota}
\end{figure}
\newcommand{\uncertaintym}{\begin{tabular}{c}Model \\ Uncertainty \end{tabular}}
\newcommand{\uncertaintya}{\begin{tabular}{c}Action \\ Uncertainty \end{tabular}}
\begin{table*}
  \caption{The results of the ablation study in terms of means and standard deviation of episode rewards over 4 seeds and last 10 trials.}
  \centering
  \begin{tabular}{ccc|cccc} \bhline{1pt}
    \multicolumn{3}{c|}{\textbf{Method}} & \multicolumn{4}{c}{\textbf{Task}} \\\hline
    & \uncertaintym & \uncertaintya & Ball-in-cup & Finger & Cheetah & Walker \\\hline
                    & $\checkmark$ & $\checkmark$  & $\underline{\mathbf{842 \pm 187}}$ & $\underline{\mathbf{818 \pm 54}}$ & $\underline{\mathbf{1738\pm 300}}$ & $\underline{\mathbf{1099\pm 153}}$ \\
    PlaNet-Bayes (Ours) & $\checkmark$ & -             & $\underline{797 \pm 219}$ & $\underline{557 \pm 296}$ & $\underline{1577\pm 329}$ & $\underline{1031\pm 167}$ \\
                    & - & $\checkmark$             & $515 \pm 431$ & $304 \pm 260$ & $1101\pm 280$ & $896\pm 99$ \\\hline
    PlaNet~\cite{hafner2018learning} & - & -       & $719 \pm 325$ & $289 \pm 284$ & $1282\pm 295$ & $845\pm 131$ \\\bhline{1pt}
  \end{tabular}
  \label{tab:ablation}
\end{table*}
\begin{figure*}[h]
  \centering
  \includegraphics[width=0.95\textwidth]{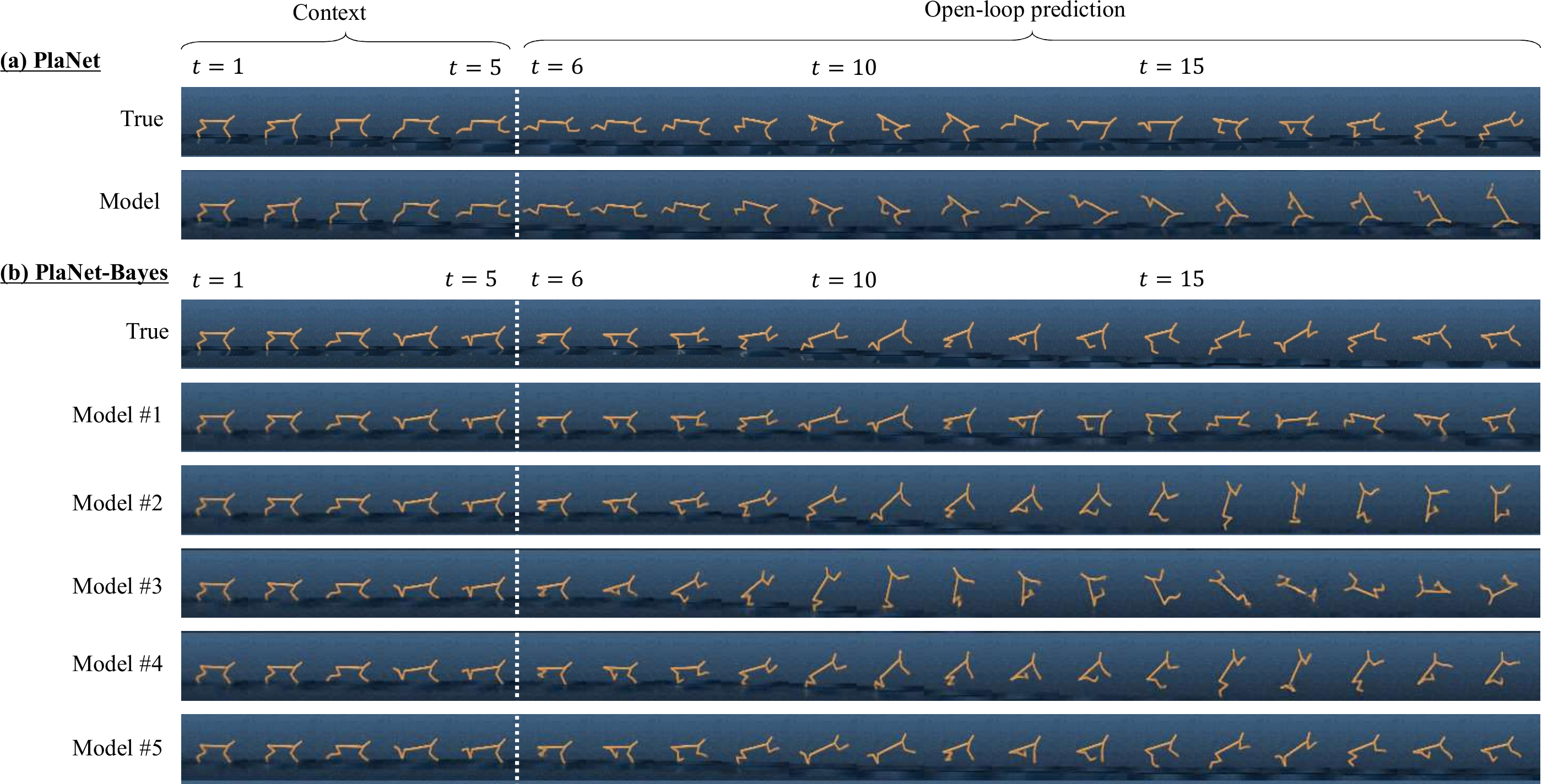}
  \caption{Open-loop video predictions. The images $t \leq 5$ show reconstructed context frames and the remaining images are generated open-loop.
  Even though common inputs are fed into PlaNet-Bayes' multiple models (neural network ensemble), they output various trajectories.
  }
  \label{fig:video_pred}
\end{figure*}
The main objective of this experiment is to demonstrate that PlaNet-Bayes has advantages over the baseline method PlaNet~\cite{hafner2018learning}.
To conduct this experiment, we consider the control tasks of the DeepMind control suite~\cite{deepmindcontrolsuite2018}, as shown in Fig.~\ref{fig:tasks}.
The four difficult domains also considered in PlaNet's paper \cite{hafner2018learning} are selected.
Concerning the task of the ``Walker'' domain, a more difficult task ``Run'' is introduced instead of the ``Walk'' task originally used in \cite{hafner2018learning}.
In addition, for several domains, their configurations are modified so as to make training more difficult%
\footnote{
For all domains except for the ``Ball-in-cup'', we modified the control range (i.e.,~torques) from $[-1, 1]$ to $[-3, 3]$, expanding action-state-spaces
and emphasizing uncertainties in the posteriors.
Without this, the optimal actions may often take clipped values,
i.e., $\action \in \{-\mathbf{1}, \mathbf{1}\}$, meaning that the orignal tasks require rather discrete control and not continuous one.
Further, in the ``Cheetah'' and ``Walker'' domains, we removed the upper limit of the reward functions. \label{footnote:task}
}.
Figure~\ref{fig:comp_sota} represents the experimental results.
It can be seen that PlaNet-Bayes consistently achieves better asymptotic performance and/or faster training compared with those of PlaNet.

\subsection{Ablation Study}
This experiment is conducted to analyze how the major components of PlaNet-Bayes (model and action uncertainty) contribute to the overall improvement.
Here, variants of the proposed have been prepared: either types of uncertainty are invalidated by setting either $E$ or $M$ equal to 1.
Table~\ref{tab:ablation} summarizes the results of the performed ablation study.
Considering only model uncertainty leads to an improvement in the performance compared with PlaNet,
while focusing on action uncertainty does not have any positive influence on the performance.
This is because action uncertainty is rather unimodal considered alone without multimodal model uncertainty modeled by a neural network ensemble.
However, when they are introduced together, multimodality in optimal actions is emphasized, and therefore, the potential of PaETS is fully exploited, achieving the best scores.

\subsection{Video Predictions}
Fig.~\ref{fig:video_pred} exemplifies video prediction results by PlaNet and PlaNet-Bayes,
in which PlaNet-Bayes generates diverse trajectory predictions by utilizing multiple models.
This uncertainty-aware behavior makes the MPC-policy avoid overfitting to unreliable predictions,
and encourage active exploration in state-action spaces.
In addition, PaETS makes the policy more active by exploiting various plans derived from the diverse predictions.

\section{Related Work} \label{sec:related_work}
Recently, MBRL applications to partially observable environments have been attracting great attention; namely this question was addressed in the studies~
\cite{ha2018world,lee2019stochastic,igl2018deep,han2019variational,kaiser2019model,okumura2020domain}.
In \cite{ha2018world}, the authors utilized a general VAE objective to train the latent space models and then optimized linear policies by using CMA-ES~\cite{hansen2003reducing}.
The time-series objective similar to Eq.~\ref{eqn:vae_elbo} was introduced in~\cite{lee2019stochastic,han2019variational},
and the learned models were applied to policy optimization by using Soft Actor-Critic~\cite{haarnoja2018soft}.
In \cite{okumura2020domain} applies PlaNet perform to imitation learning by employing adversarial training.
Although most of these studies were based on autoencoding variational Bayes,
posterior inference of models was not considered.
In \cite{tran2019bayesian}, there was an attempt to extend PlaNet to be Bayesian by employing variational inference of the posterior $\modelpost$. However, as experimentally suggested in \cite{okada2019variational},
a tractable variational distribution $q(\theta)$, i.e., Gaussian, is less expressive than the neural network ensemble to capture multimodality.
In addition, the experiment presented in \cite{tran2019bayesian}, which was conducted using the ``Cheetah'' domain, did not the achieve asymptotic performance improvement.
Regarding uncertainty-aware planning, the research works mentioned above did not consider the action uncertainty.
To the best of our knowledge, PlaNet-Bayes is the first attempt to incorporate uncertainty and demonstrate its efficiency by achieving better performance compared with the state-of-the-art latent MPC method: PlaNet.


\section{Conclusion} \label{sec:conclusion}
In the present paper, we proposed PlaNet-Bayes, a Bayesian extension of the state-of-the-art MBRL method for partially observable environments PlaNet.
Bayesian inference of the model posterior $\modelpost$ was introduced
such as to be realized by the ensemble of latent dynamics models.
The other Bayesian concept was employed for inference of the optimal trajectory posterior, which allowed formulating VI-MPC and PaETS for latent planning to successfully incorporate the multimodal uncertainty in optimal actions.
By exploiting both considered uncertainty-aware strategies, PlaNet-Bayes was able to improve asymptotic performance compared to with that of baseline PlaNet.

The approaches proposed in this paper are also applicable to a variety of control methods for partially observable environments.
For example, the ensemble scheme of the latent state-space model could be used to improve policy-oriented MBRL performance~\cite{ha2018world,lee2019stochastic,han2019variational}.
On the other hand, by introducing a categorical distribution (or a mixture model of it) as a variational distribution $q$, latent planning for discrete control tasks, including video- and board-games~\cite{kaiser2019model,silver2016mastering,schrittwieser2019mastering}, is possible.

Future research directions will be related to experiments on latent MPC in real systems.
Specifically in the case of industrial systems, deep understanding of the learned models and latent space is essential to obtain information about the agent's decision-making process.
To represent the explainable latent space, we will consider adopting  disentangling approaches~\cite{thomas2018disentangling,sawada2018disentangling} as a promising research direction.
The proposed model ensemble approach may also contribute to analysis and enhancement of the decision-making, as the confidence of trajectory prediction can be quantified by using multiple prediction outputs similarly as in~\cite{beluch2018power},
which suggests what the new data can be collected to improve the confidence by reducing the uncertainty.

\bibliography{iros2020}
\bibliographystyle{ieeetr}

\begin{appendices}

\section{Complete Definition of PaETS} \label{sec:paets_eqs}

\begin{align}
    w_{k}^{(j+1)} &\leftarrow
    \frac
    {\mathcal{W}(\action_{k})^{\frac{1}{\lambda}}\cdot (q^{(j)}(\action_{k}))^{-\kappa}}
    {\displaystyle \sum_{k'=1}^{K}\mathcal{W}(\action_{k'})^{\frac{1}{\lambda}}\cdot (q^{(j)}(\action_{k'}))^{-\kappa}} \\
    \resp_{m}(\action_{k}) &\coloneqq \frac
    {\mixcoef^{(j)}_{m}\mathcal{N}(\action_{k} ; \actionm^{(j)}_{m}, \actions^{(j)}_{m})} 
    {\displaystyle \sum^{M}_{m'=1}\mixcoef^{(j)}_{m'}\mathcal{N}(\action_{k} ; \actionm^{(j)}_{m'}, \actions^{(j)}_{m'})} \label{eqn:resp} \\
    \omega_{m, k}^{(j+1)} &\coloneqq 
        {\resp_{m}(\action_{k}) w_{k}^{(j+1)}} {\Big /}
        {\underbrace{\sum^{K}_{k'=1} \resp_{m}(\action_{k'}) w_{k'}^{(j+1)}}_{\coloneqq N_{m}}} \\
    \actionm_{m}^{(j+1)} &\leftarrow
        {\sum^{K}_{k=1} \omega^{(j+1)}_{m,k}\action_{k}} \\
    \actions_{m}^{(j+1)} &\leftarrow
        {\sum^{K}_{k=1} \omega^{(j+1)}_{m,k}(\action_{k} - \actionm_{m}^{(j+1)})^{2}} \\
    \mixcoef_{m}^{(j+1)} &\leftarrow N_{m} {\Big /} \sum^{M}_{m'=1} N_{m'}.
\end{align}

\end{appendices}

\end{document}